\documentclass[conference]{IEEEtran}
\IEEEoverridecommandlockouts
\usepackage{amsmath,amssymb,amsfonts}
\usepackage{algorithmic}
\usepackage{graphicx}
\usepackage{textcomp}
\usepackage{xcolor}
\usepackage{caption}
\usepackage{setspace, lipsum}
\usepackage{hyperref}
\usepackage{multirow}
\usepackage{color, colortbl}
\usepackage{url}
\usepackage{subcaption}
\usepackage{float}
\usepackage{nccmath}
\usepackage{amsthm,bbm}
\usepackage{amsfonts}
\usepackage{adjustbox}
\usepackage{pifont}
\hypersetup{
    colorlinks,
    linkcolor={red!50!black},
    citecolor={blue!50!black},
    urlcolor={blue!80!black}
}

\definecolor{darkred}{rgb}{0.7, 0, 0}
\definecolor{darkblue}{rgb}{0.0, 0.0, 0.7}
\definecolor{darkgreen}{rgb}{0, 0.4, 0}
\newcommand{\cmark}{\textcolor{darkgreen}{\ding{51}}}
\newcommand{\xmark}{\textcolor{darkred}{\ding{55}}}

\usepackage[sorting = none, backend = bibtex, style=numeric-comp]{biblatex}
\AtBeginBibliography{\small}

\begin{document}

\title{Urban-StyleGAN: Learning to Generate and Manipulate Images of Urban Scenes \\

\thanks{\textsuperscript{*} Equal Contribution}
\thanks{ Corresponding author email: george.eskandar@iss.uni-stuttgart.de}
\thanks{The research leading to these results is funded by the German Federal Ministry for Economic Affairs and Energy within the project "AI Delta Learning." The authors would like to thank the consortium for the successful cooperation.}
\thanks{Code is available at \url{https://github.com/GeorgeEskandar/UrbanStyleGAN}}
}

\author{\IEEEauthorblockN{George Eskandar\textsuperscript{1}\textsuperscript{*}, Youssef Farag\textsuperscript{2}\textsuperscript{*}, Tarun Yenamandra~\textsuperscript{2},  Daniel Cremers~\textsuperscript{2}, Karim Guirguis\textsuperscript{3}, Bin Yang\textsuperscript{1}}
\IEEEauthorblockA{\textsuperscript{1} University of Stuttgart, Germany }
\IEEEauthorblockA{\textsuperscript{2} Technical University of Munich (TUM), Germany}
\IEEEauthorblockA{\textsuperscript{3} Bosch Center for Artificial Intelligence, Renningen, Germany}
}

\maketitle

\begin{abstract}

A promise of Generative Adversarial Networks (GANs) is to provide cheap photorealistic data for training and validating AI models in autonomous driving. Despite their huge success, their performance on complex images featuring multiple objects is understudied. While some frameworks produce high-quality street scenes with little to no control over the image content, others offer more control at the expense of high-quality generation. A common limitation of both approaches is the use of global latent codes for the whole image, which hinders the learning of independent object distributions. Motivated by SemanticStyleGAN (SSG), a recent work on latent space disentanglement in human face generation, we propose a novel framework, Urban-StyleGAN, for urban scene generation and manipulation. We find that a straightforward application of SSG leads to poor results because urban scenes are more complex than human faces. To provide a more compact yet disentangled latent representation, we develop a class grouping strategy wherein individual classes are grouped into super-classes. Moreover, we employ an unsupervised latent exploration algorithm in the $\mathcal{S}$-space of the generator and show that it is more efficient than the conventional $\mathcal{W}^{+}$-space in controlling the image content. Results on the Cityscapes and Mapillary datasets show the proposed approach achieves significantly more controllability and improved image quality than previous approaches on urban scenes and is on par with general-purpose non-controllable generative models (like StyleGAN2) in terms of quality. 

\end{abstract}


\section{Introduction}
\label{sec:intro}
Synthetic data produced by graphics engines or driving simulators has become prevalent~\cite{carla, gtav} because it can model new sensors and driving situations providing large amounts of training and validation data to autonomous driving AI models. However, synthetic images are not sufficiently photorealistic, and models trained on synthetic data alone do not readily generalize to real data. Utilizing Generative Adversarial Networks (GANs)~\cite{Goodfellow2014GenerativeAN} is a possible alternate approach, as they have demonstrated huge success, achieving more photorealism than computer graphics in some applications like human face generation~\cite{stylegan}. 


\begin{figure}[t!]
    \centering
    \begin{center}
    \includegraphics[width=0.8\linewidth]{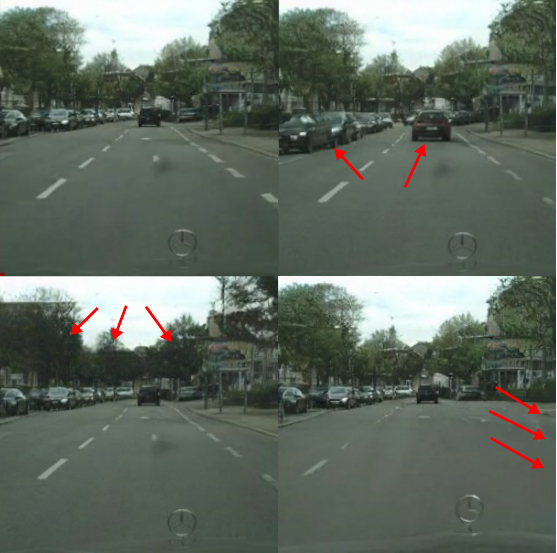}
    \end{center}
    \caption{ \small Our framework can generate and/or edit high-quality images. An example of a generated scene from our model is shown in the top left. After generation, we can edit the image in multiple different ways: increase the car size in the scene (Top Right), have trees with fewer leaves (bottom left), or have a wider road (bottom right).}
    \label{fig:teaser}
    \vspace{-1.5em}
\end{figure}
The success of GANs has mostly been achieved on images with single objects and is yet to reach multi-object complex scenes. While some GAN models can generate high-quality street-level scenes like the StyleGAN series~\cite{stylegan, stylegan2, stylegan3, styleganxl} and ProjectedGAN~\cite{projected}, they do not provide any level of controllability for localized edits in the image. Other models provide slightly more controllability, but their quality lags behind. Most importantly, SB-GAN~\cite{sbgan} used a two-stage approach to generate complex images by first generating a semantic layout from a latent code, then using the layout as a prior to generate an image with a conditional GAN like SPADE~\cite{spade}. However, image manipulation can only be achieved by manually editing the generated layout from the first stage. Semantic Palette~\cite{palette} has built upon this approach by providing the label generator with a conditional input of class distributions ($40\%$ sky, $10\%$ cars...). However, it still does not reach the quality of StyleGAN and provides a small level of controllability over the generation, as specifying the class statistics only does not offer fine-grained edits like changing the size, number, and/or location of a class in the generated image. For instance, increasing the percentage of the class 'cars' in an image can be achieved either by increasing the number or size of cars in the scene, and both can happen randomly as this is not controlled in Semantic Palette. 

In this work, we aim to develop a generative framework that simultaneously achieves a high level of photorealism and controllability. Controllability is highly desired for autonomous driving applications, especially for model validation, because it enables counterfactual reasoning: for example, we wish to know how a perception model would behave had there been more cars on the road. To this end, we propose a novel approach, Urban-StyleGAN, to learn the generation and manipulation of urban scenes (see Figure~\ref{fig:teaser}). Contrary to previous approaches, which only use latent codes for the global image, Urban-StyleGAN is based on the idea that a disentangled latent code allows localized edits in the image. Specifically, we first adopt a recent work from human face generation, namely, SemanticStyleGAN~\cite{ssg} (SSG) but find that a straightforward application leads to training divergence. We hypothesize that, the large number of classes in urban scenes increases the number of local generators, and, consequently, the generator's overall learning capacity and the latent space dimension. This imposes a more complicated learning problem. As a remedy, we propose a class grouping strategy to learn super-classes, effectively reducing the number of local generators and speeding up the training convergence. Moreover, to allow localized semantic edits, we employ recent unsupervised latent exploration algorithms on the $\mathcal{S}$-space of the disentangled class codes of SSG, which contrasts the common use of these approaches on the $\mathcal{W}^+$-space. Our contributions can be summarized as follows: 
\begin{itemize}
    \item We propose a framework that can simultaneously generate high-quality images of urban scenes and enables fine-grained control over the image post-synthesis. Key to our method is a pre-training class grouping strategy for limiting the number of local generators and, as a result, the generator's total learning capacity. This allows a better exploitation of SSG's disentangled latent space. 
    \item To promote more controllability on the image content, we employ Principal Component Analysis (PCA) in the lower dimensional disentangled $\mathcal{S}$-space for each class rather than the $\mathcal{W}^+$-space, which has often been used in previous publications.  To the best of our knowledge, this is the first paper to explore directions of control in the latent space of a GAN trained on urban scenes.  
    \item Experiments on Cityscapes and Mapillary datasets show that the proposed model offers fine-grained localized semantic edits (e.g., the number, size, and position of objects in a scene) and outperforms previous urban scene generative models by a large margin in generation quality and controllability.
\end{itemize}
 
\section{Related Works}
\label{sec:rw}
\noindent\textbf{Generative Adversarial Networks (GANs)} have revolutionized image generation by employing an adversarial learning scheme between a generator and a discriminator network~\cite{Goodfellow2014GenerativeAN}. Since then, a plethora of works\cite{Karras2018ProgressiveGO, stylegan, stylegan2, stylegan3, styleganxl, projected, fastgan} has progressively enhanced the diversity and quality of generated images. While a high-quality synthesis can be achieved on urban scenes, the latent codes of these frameworks affect the image globally and do not provide mechanisms for localized edits. 

\noindent\textbf{Semantic prior for image generation.} Two recent works have emerged that tackle urban scene generation: SB-GAN~\cite{sbgan} and Semantic Palette~\cite{palette}. A two-stage approach is used in both models: the first stage generates semantic layouts, and the second stage translates the layouts into images, using a conditional network like~\cite{spade, oasis, ccfpse, pix2pixhd, usiscag}. While they provide more controllability than general-purpose GANs (like StyleGAN), they lag in generation quality. We hypothesize that a limiting factor in both frameworks~\cite{sbgan, palette} is the conditional network used in the second stage and argue that the joint generation of images and layouts can boost performance. Most recently, SemanticStyleGAN~\cite{ssg} (SSG) proposed to generate images and layouts in one stage, with end-to-end training, but it was designed and applied for human face generation. 

\noindent\textbf{Image manipulation in GANs} has been enabled in 3 different ways: using synthetic data to provide additional supervision~\cite{discofacegan, config, gancontrol}, exploring directions of control in the latent space in an unsupervised manner~\cite{ganspace, interfacegan, styleflow, sefa}, and using language prompts~\cite{stylemc, styleclip}. These methods have been applied to single-object datasets, we found no previous works on exploring latent space for generative models on urban scenes. Our work belongs to the last category (unsupervised latent exploration) and is based on GANSpace~\cite{ganspace}. We opt not to adopt any synthetic datasets for additional supervision in order to generate photorealistic data only. We leave the third method (language prompts) for future works.
 
\section{Methodology}
\label{sec:method}
\begin{figure*}[t]
    \centering
    \begin{center}
    \includegraphics[width=0.9\textwidth]{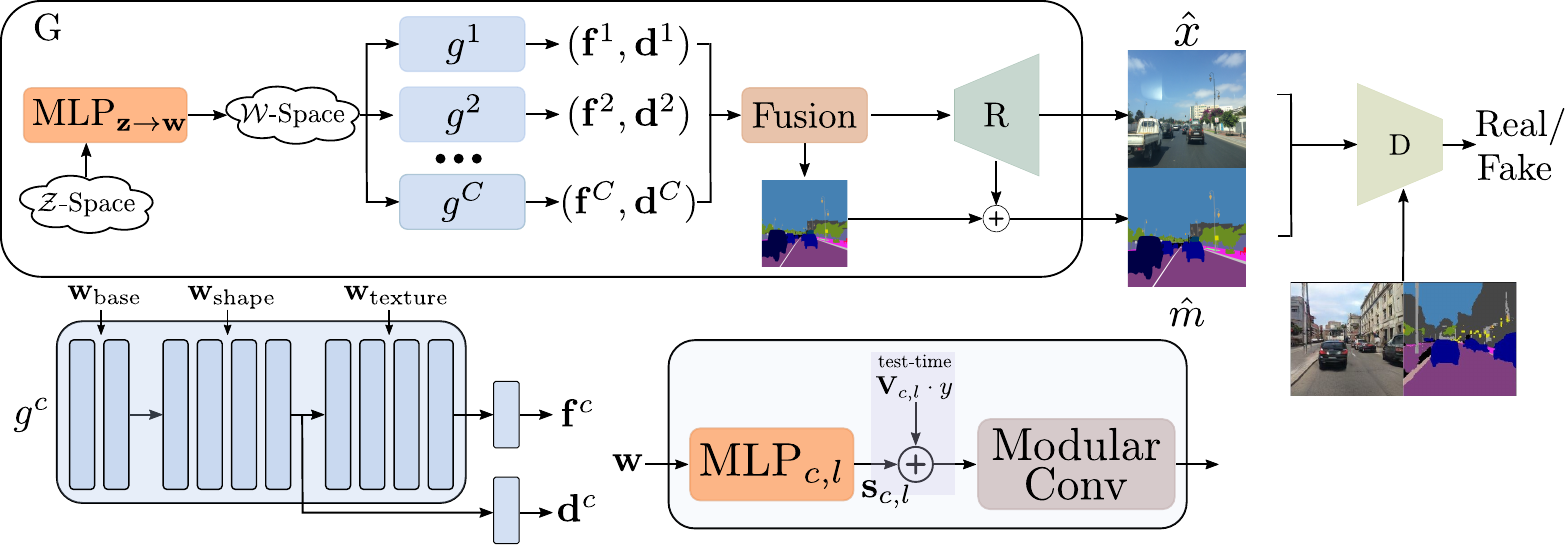}
    \end{center}
    \caption{Architecture of the proposed framework, Urban-StyleGAN. The number of local generators is controlled by the number of super-classes defined by the class-remapping table. In test-time, we apply GANSPace on the disentangled latent $\mathbf{s}_{c,l}$-vectors to discover meaningful directions of control.    }
    \label{fig:framework}
    \vspace{-1em}
\end{figure*}
Our key idea is to build a framework with a disentangled latent code to learn the different factors of variation in an urban scene. For this, we adopt a recent baseline on human face generation, SSG~\cite{ssg}. SSG learns to generate semantic layouts using different local generators to model the different classes in the image. Similar to SB-GAN and Semantic Palette in urban scene generation, the learning of semantic labels is used as a prior for generating images and learning the shape of different classes, but SSG learns both images and layouts in an end-to-end manner instead of using a two-stage approach. We argue that an end-to-end approach can lead to a faster training convergence as the generator learns shape and texture jointly. However, we show that a straightforward application of SSG does not lead to convergence or high quality. As a remedy, we propose Urban-StyleGAN to generate and edit high-quality images of urban scenes. This is realized by learning classes in groups, which enables a more compact representation in the latent space and a faster training convergence. Moreover, after training, we choose to explore and find the main directions of variation in the class-specific $\mathcal{S}$-space, instead of the $\mathcal{W}^+$-space, to allow for more localized semantic edits.    
\subsection{Overview of the framework}
The StyleGAN series of models~\cite{stylegan, stylegan2, stylegan3, styleganxl} is based on the idea of injecting multiple latent codes in different generator layers to modulate \textit{conv} layers at different scales. To obtain these latent codes, a multilayer perceptron (MLP) maps vectors from a normal distribution called the $\mathcal{Z}$-space to vectors $\mathbf{w}$ in an intermediate space called $\mathcal{W}$-space. As the $\mathbf{w}$ vectors affect the image globally, localized edits are not possible. In this paper, we challenge this paradigm and draw inspiration from SSG~\cite{ssg} for the design of the generator's architecture. We employ multiple local generators $g^c$, where each $g^c$ maps a latent code $\mathbf{w}$ and Fourier features $\mathbf{p}$~\cite{stylegan3} to a class-specific depth map $\mathbf{d}^c$ and class-specific features $\mathbf{f}^c$ for a class $c$ in the image. The features $\{\mathbf{f}^c\}_{c=1}^C$ are then fused in a compositional manner resulting in an intermediary semantic mask $\mathbf{m}$ and features $\mathbf{f}$:
{
 \begin{align}
\mathbf{f}^c, \hfill \mathbf{d}^c = g^c(\mathbf{w}, \mathbf{p}), \quad \mathbf{m}^c_{i,j} = \frac{e^{\mathbf{d}^c_{i,j}}}{\sum_{k=1}^C e^{\mathbf{d}^k_{i,j}}}, \quad f = \sum_{k=1}^C \mathbf{m}^k \cdot \mathbf{f}^k.
 \label{eq:composition}
 \end{align}
}
A renderer network $R$ upsamples $\mathbf{f}$ to a high-resolution image $\hat{\mathbf{x}}$ and a final semantic mask $\hat{\mathbf{m}}$ (see Figure \ref{fig:framework}). For this, the generator ($R$ and $\{g^c\}_{c=1}^C$) trains adversarially on a labeled dataset with images and semantic layouts. The discriminator architecture has two branches for the image and mask, which are fused early in the network. A realism score is given at the end using a fully convolutional layer (for more details about the discriminator's architecture, readers are referred to SSG~\cite{ssg}).   

The $\mathbf{w}$ vectors are factorized in three components: $\mathbf{w} = (\mathbf{w}_{\mathrm{base}}, \mathbf{w}_{\mathrm{shape}}, \mathbf{w}_{\mathrm{texture}})$, where each modulates different layers in $g^c$. $\mathbf{w}_{\mathrm{base}}$ determines the coarse-level shape of the semantic class and is shared across all classes to model their spatial dependencies. The vector $\mathbf{w}_{\mathrm{shape}}$ determines the class shape and $\mathbf{w}_{\mathrm{texture}}$ its texture. This is enforced by architectural design, as the depth map, which determines the mask's shape, only depends on $\mathbf{w}_{\mathrm{base}}$ and $\mathbf{w}_{\mathrm{shape}}$ (Figure~\ref{fig:framework}). Each $\mathbf{w}$-subvector (base, shape, texture) is passed to a set of layers inside each generator $g^c$ and transformed to a style vector, $\mathbf{s}_{c,l}$, through an MLP inside the layer $l$ of $g^c$ (see Figure \ref{fig:framework}). The style vectors modulate the \textit{conv} layers inside each generator. Formally, the mapping from $\mathbf{z}$ to $\mathbf{s}_{c,l}$ can be denoted as follows: 

{
 \begin{align}
&\mathrm{MLP}_{\mathbf{z} \rightarrow \mathbf{w}}: \mathbf{z} \longrightarrow \mathbf{w} = (\mathbf{w}_{\mathrm{base}}, \mathbf{w}_{\mathrm{shape}}, \mathbf{w}_{\mathrm{texture}}),\\
&\mathrm{MLP}_{c, l}: \mathbf{w}_{\mathrm{base}} \longrightarrow \mathbf{s}_{c,l=0,1}, \mathbf{w}_{\mathrm{shape}} \longrightarrow \mathbf{s}_{c,l=2:5}, \\
&\quad \quad \quad \quad \mathbf{w}_{\mathrm{texture}} \longrightarrow \mathbf{s}_{c,l=6:9}.
 \end{align}
}

Note that during training, we sample one $\mathbf{w}$-vector for all classes, while during inference, we can sample different $\mathbf{w}$ for different classes. The ensemble of all $\mathbf{w}$ vectors is called the $\mathcal{W}^+$-space and the ensemble of all $\mathbf{s}$-vectors in the generator is called the $\mathcal{S}-space$ and contains all vectors $\{\mathbf{s}_{c,l}\}_{c=1, l=1}^{C, L}$.

\begin{table}
\centering

\adjustbox{width=0.9\linewidth}{\begin{tabular}{ |c|c| }
 \hline
 \textbf{Super-Class} & \textbf{Original Classes} \\
 \hline
 Void & Classes 0-6 in Cityscapes\\
 \hline
 Drive-able & Road, Parking, Rail Track \\
 \hline
 Side Walk& Side Walk \\
 \hline
 Building & Building, Bridge, Tunnel \\
\hline
 Wall & Wall, Guard Rail\\
 \hline
 Fence & Fence \\
\hline 
Person & Person \\
\hline 
Car & Car \\
\hline
Other Vehicles & Bus, Train, Truck, Caravan, Trailer \\
\hline
Bike & Bicycle, Motorcycle \\
\hline 
Rider & Rider \\
\hline 
Sky & Sky \\
\hline 
Greenery & Terrain, Vegetation \\
\hline 
Traffic Light & Traffic Light \\
\hline 
Traffic Sign & Traffic Sign \\
\hline 
Poles & Poles, Poles Group \\
\hline
\end{tabular}}
\caption{Mapping table from 34 classes to 16 super-classes in the Cityscapes dataset. This is executed as a preprocessing step on the semantic layouts.}
\label{table:cityscapes_mapping_1}
\vspace{-1.5em}
\end{table}
\subsection{Learning high-quality generation of urban scenes}
We find that this architecture, originally designed for human face generation, does not readily extend to the more complex street-level images. As autonomous driving datasets typically contain a high number of classes (30-60 compared to 8 in human faces datasets), the number of local generators increases significantly. This, in turn, leads to 2 problems. First, the generator has a larger learning capacity and overpowers the discriminator, introducing overfitting (see Figure~\ref{fig:fid_plot}). Second, as some classes are less frequent in the dataset, allocating a local generator to learn each class makes the mapping task from latent code to feature substantially harder because the latent space has to be more descriptive. To mitigate these problems, we propose to remap the original classes into a smaller number of super-classes, so they can be learned in groups. This reduces the number of local generators to the number of super-classes and consequently reduces the generator's overall learning capacity. Moreover, it enables a compact yet disentangled latent representation of the objects in the scene. However, we do not want to reduce the number of classes drastically in a way that impedes controllability and disentanglement, and we find that 16 super-classes is a good balance (see Table~\ref{table:cityscapes_mapping_1} for an example of grouping on the Cityscapes dataset~\cite{cordts2016cityscapes}). To further regularize the adversarial training, we add spectral normalization~\cite{spectralnorm} to the discriminator weights, except for the last fully-connected layers, and increase the batchsize in the training pipeline. While spectral normalization regularizes large changes in the weights leading to more stable updates, a larger batchsize fosters the discriminator's training as it can observe all of the 16 classes in one forward pass. Otherwise, the discriminator learns to neglect some classes, especially the less frequent ones in the dataset.

\subsection{Exploiting the disentangled representation of super-classes}
The architecture of SSG enables to have control over the shape and texture of different classes but does not offer by design any clear directions of control. Simply increasing or decreasing the latent code can have multiple simultaneous effects on a class shape but we wish to find fine-grained directions (like only changing the number of cars on the right or changing the time of day...). To this end, explore latent directions in an unsupervised way like~\cite{ganspace}, as we do not have additional supervision from synthetic data like~\cite{gancontrol} or attribute classifier networks like~\cite{styleflow}. Specifically, we sample $N$ vectors from the normal distribution in the $\mathcal{Z}$-space and compute the corresponding class-specific style vectors $\mathbf{s}_{c,l=5,9}^{1:N}$, where $l=5,9$ for shape and texture changes respectively. Then, to find the principal axes of the probability distribution of the $\mathbf{s}$-vectors, we perform PCA on each subset of vectors, $\mathbf{s}_{c,l}^{1:N}$ and obtain two bases per class, $\mathbf{V}_{c,l=5}$ for shape editing and $\mathbf{V}_{c,l=9}$ for texture editing. Note that we only choose these 2 layers, $l=5,9$, inside each local generator to perform shape or texture changes. The layer $l=5$ corresponds to the last layer in the subnetwork of $g^c$ responsible for the generation of $\mathbf{d}^c$, while $l=9$ is the last layer which outputs the feature map $\mathbf{f}^c$. We found that these downstream layers provide more meaningful directions of control than the earlier layers (Figure~\ref{fig:tab2}). A possible explanation is that they operate on a larger receptive field of view, generating more high-level features.

Formally, editing can be expressed as changing the $\mathbf{s}$-vector in the desired direction(s) to a new vector $\mathbf{s}_{c,l}^{'}$: 

{
 \begin{align}
\mathbf{s}_{c,l}^{'} = \mathbf{s}_{c,l} + \mathbf{V}_{c,l} \cdot y, 
 \end{align}
}
where $y$ is a vector of real-valued coordinates, which determines the specified change by the user in the desired direction. Note that in contrast to the original GANSpace algorithm~\cite{ganspace}, we do not apply PCA in the $\mathcal{W}^+$-space. Instead, we seek to edit the image by manipulating the $\mathbf{s}$-vector. Our motivation for this is that the $\mathbf{s}_{c,l}$ are more \textit{class-specific} and \textit{layer-specific} than the $\mathbf{w}$-vectors, as they are generated by different $\mathrm{MLP}_{c,l}$ located inside the different generators $g^c$, whereas the $\mathbf{w}$-vectors are originated from $\mathbf{z}$ through one layer, $\mathrm{MLP}_{\mathrm{shared}}$ and the $\mathcal{W}$-space is shared across all classes. 

\section{Experiments}
\label{sec:experiments}
\noindent\textbf{Datasets.} We use two datasets for our experiments, Cityscapes~\cite{cordts2016cityscapes} and Mapillary Vistas~\cite{mapillary}.  Cityscapes contains street-level images from different cities in Germany, annotated with $34$ semantic classes. The original training split has $3\mathrm{k}$ annotated images and $20\mathrm{k}$ images without annotations. Similar to Semantic Palette~\cite{palette}, we employ a segmentation network to generate annotations for the unlabeled images and use the $23\mathrm{k}$ images and labels for training. Mapillary Vistas is larger and more diverse than Cityscapes and contains $25\mathrm{k}$ annotated images with $66$ semantic classes in diverse cities worldwide. All images are generated with a resolution of $256 \times 256$. 

\noindent\textbf{Baselines and Metrics.} In terms of generation quality, we compare with general-purpose generative models like ProGAN~\cite{Karras2018ProgressiveGO}, StyleGAN2~\cite{stylegan2}, VQGAN~\cite{vqgan}, ProjectedGANs~\cite{projected}, SAGAN~\cite{sagan}, and generative models for urban scenes namely: Semantic Palette~\cite{palette} and SB-GAN~\cite{sbgan}. Following StyleGAN2, we measure the Frechet Inception Distance (FID) between $50\mathrm{k}$ generated images and all images from the corresponding real dataset. Note that the last two baselines~\cite{palette, sbgan} generate label maps and images like our approach, so we also measure the mIoU between generated images and generated layouts using pre-trained segmentation networks on the real datasets~\cite{segmentor}. Note that after performing the class grouping strategy, the semantic layouts contain a different number of classes. For a fair comparison, we train SP and SB-GAN on the same layouts and measure the FID and mIoU.  

\noindent\textbf{Baselines to assess controllability.} In terms of controllability, we compare with SP and the original SSG baseline (after introducing our changes to generate high-quality images) in Figure~\ref{fig:tab1}. SP offers controllability by changing either the $\mathbf{z}$-vector or a condition vector containing the class distributions. For example, to generate more buildings, we would change this vector to increase the percentage of buildings and reduce the percentage of trees in the image. SSG native controllability is to sample different $\mathbf{w}$-vectors for different classes and increase/decrease the value of these vectors with a constant parameter. In Figure~\ref{fig:tab2}, we also explore applying GANSpace in the $\mathcal{W}^+$-space of Urban-StyleGAN and in the $\mathcal{S}$-space of different layers.

\section{Results}
\label{sec:results}
\subsection{Generation results}
\noindent\textbf{Main comparisons.} In Table~\ref{table:city_benchmark}, we compare with general-purpose generative models on the Cityscapes benchmark. The proposed approach outperforms most previous methods and is almost on par with StyleGANv2 and only inferior to ProjectedGAN~\cite{projected}. However, both ProjectedGAN and StyleGAN2 do not provide controllability over the image generation as they use global latent codes. On the other hand, Urban-StyleGAN largely outperforms approaches that provide controllability through the use of a semantic prior~\cite{sbgan, palette} on both metrics (FID and mIoU). In this work, our primary objective is not to achieve the highest generative quality but to achieve high controllability in the generation process with strong quality and fidelity metrics. 


\begin{table}[t!] 
\centering  
\begin{tabular}{ccc} 
\hline
Method & FID $\downarrow$ & mIoU $\uparrow$ \\
\hline  
ProGANs~\cite{Karras2018ProgressiveGO} & 63.87 & - \\
VQGAN~\cite{vqgan} & 173.80 & - \\
SB-GAN~\cite{sbgan} & 62.97 & 30.04 \\
SAGAN~\cite{sagan} & 12.81 & - \\
SP~\cite{palette} & 52.5 & 21.40 \\
StyleGAN2~\cite{stylegan2} & 8.35 & - \\
ProjectedGANs~\cite{projected} & 3.41 & - \\
\textbf{Ours} & 9.8 & 34.0\\
\hline
\end{tabular}

\caption{Image Synthesis on Cityscapes, Resolution $256^2$.} \label{table:city_benchmark} 
\end{table}
\begin{table}[t!] \centering  
\begin{tabular}{cccc} 
\hline
Super-classes & SpectralNorm & Batchsize & FID $\downarrow$ \\
\hline
34 & \xmark & 4 & 280.5\\
16 & \xmark & 4 & 36.6 \\
16 & \cmark & 4 & 23.8 \\
\hline 
16 & \cmark & 4 & 23.8 \\
16 & \cmark & 12 & 21.4\\
16 & \cmark & 16 & 9.8 \\
\hline
24 & \cmark & 12 &  38.3 \\
16 & \cmark & 12 & 21.4 \\
12 & \cmark & 12 & 13.5 \\
9 & \cmark  & 12 & 11.4 \\
\hline
16 & \cmark & 16 & 9.8 \\

\hline
\end{tabular}

\caption{Ablation Study on Cityscapes, Resolution $256^2$.} \label{table:ablation} \vspace{-1.5em} \end{table}
\begin{figure}[t]
    \centering
    \begin{center}
    \includegraphics[width=0.95\linewidth]{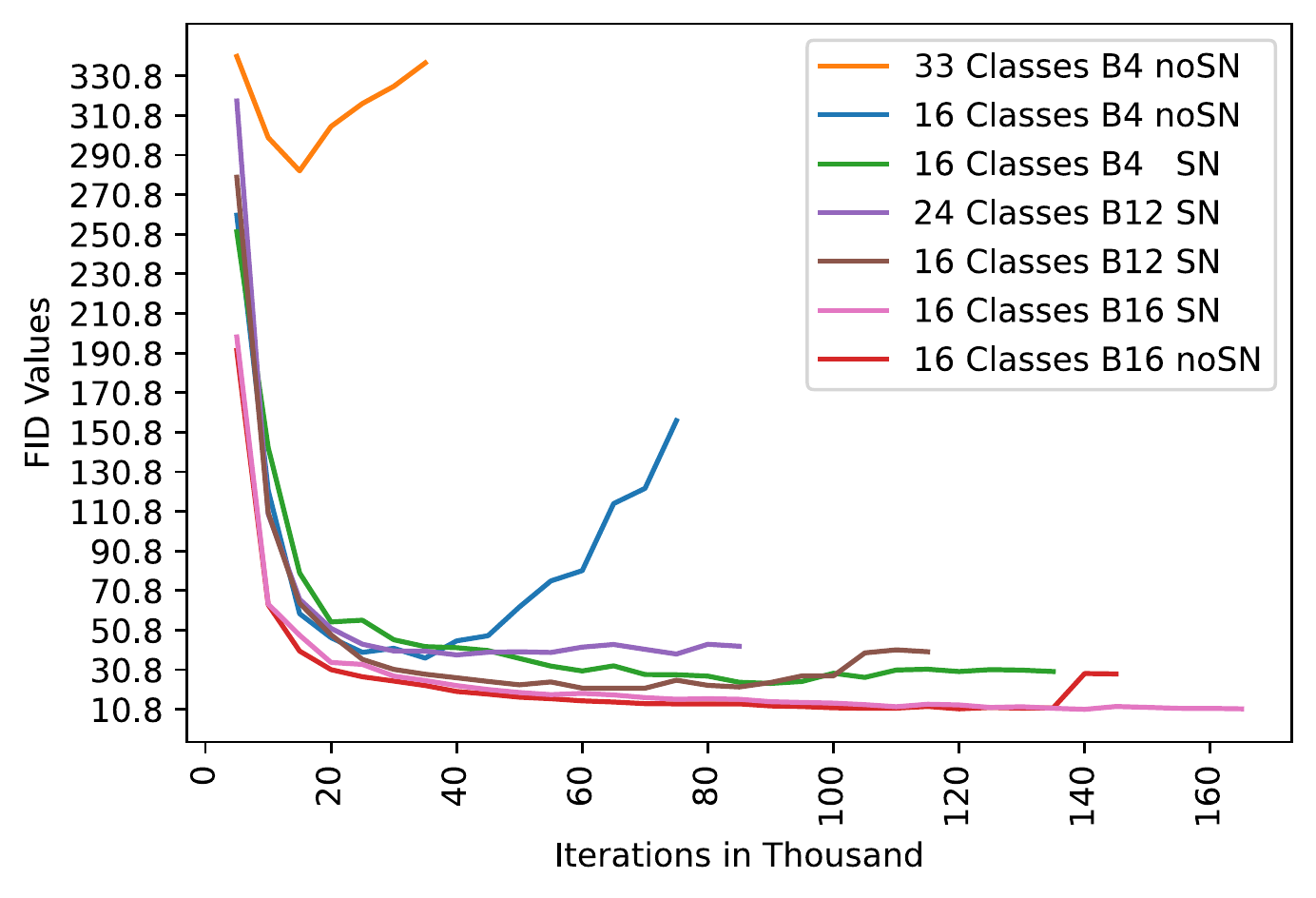}
    \end{center}
    \caption{Convergence of SSG architecture before and after the proposed modifications. SN refers to spectral normalization in the discriminator and B the batchsize.}
    \label{fig:fid_plot}
    \vspace{-1em}
\end{figure}

\noindent\textbf{Ablation on generator's architecture and training settings.} We perform ablation studies on the architecture and regularization strategy on the Cityscapes dataset and report the results in Table~\ref{table:ablation}. The FID plot for different experiments is shown in Figure~\ref{fig:fid_plot}. First, we show that running SSG with the full number of classes reaches a high FID and diverges after a short training time. Grouping the classes into 16 superclasses, we reduce the number of local generators $g^c$, providing an architectural regularization to the adversarial framework. The FID drops significantly to $36.6$, but the training still diverges after some time. Hypothesizing that the generator still overpowers the discriminator, we regularize the discriminator by applying spectral normalization on the convolutional weights, dropping the FID further to $23.8$ and leading to training convergence. We also show that a large batchsize is fundamental to reaching a better FID. Our intuition for this is that the discriminator can observe more classes in one forward pass with a larger batchsize, which leads to stronger feedback to the generator. We also experiment with a different number of super-classes (9,12,16,24) and notice that it is inversely proportional to the FID. Note that reducing the number of super-classes to a certain point would compromise controllability. We reach the best performance with $16$ classes when we increase the batchsize to $16$ and use spectral normalization. Note that we choose to have $16$ classes (instead of $9$ or $12$) to allow for more controllability.

\subsection{Controllability results}
\begin{figure*}[t]
    \centering
    \begin{center}
    \includegraphics[width=0.9\textwidth]{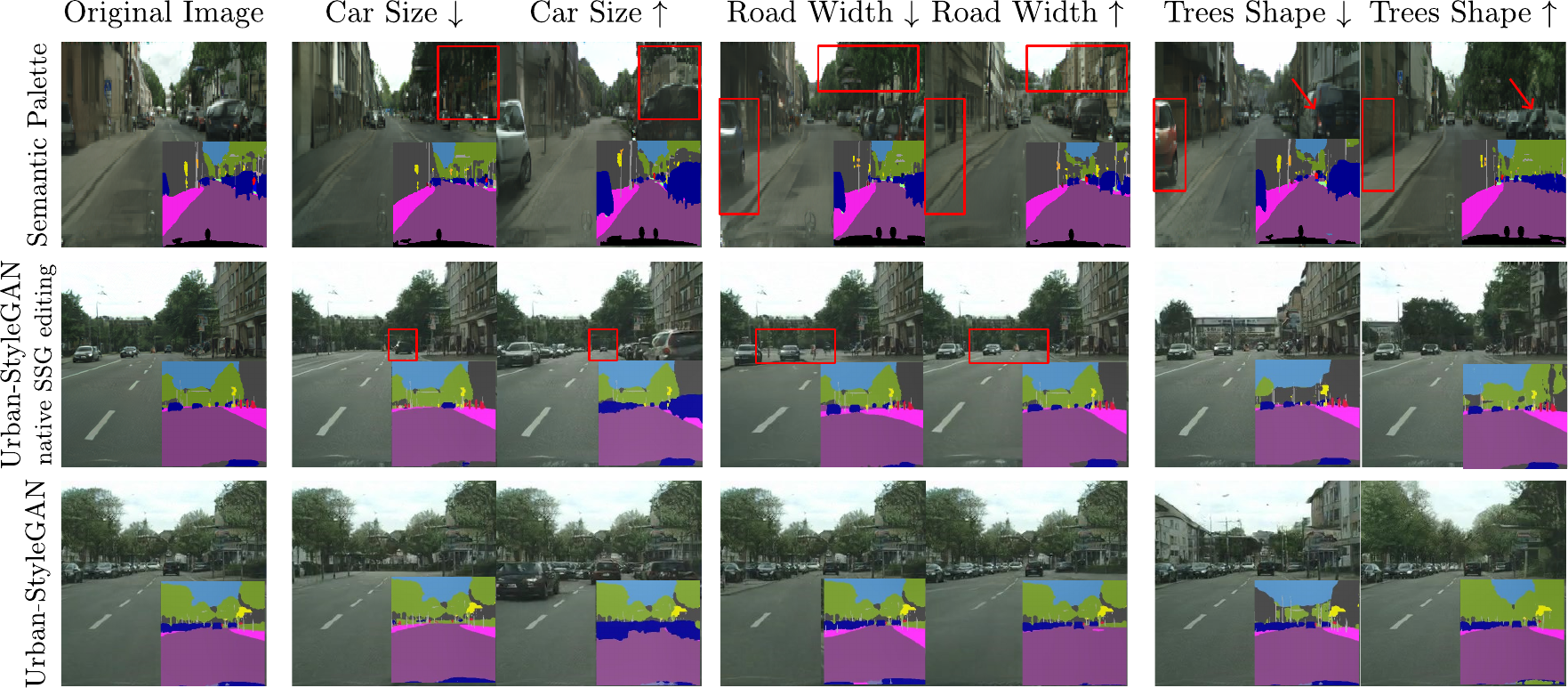}
    \end{center}
    \caption{ Single Class manipulation on Cityscapes. Generated Images and labels are shown. Red Boxes denote inconsistencies between manipulated images. }
    \label{fig:tab1}
    \vspace{-1em}
\end{figure*}
\begin{figure}[h!]
    \centering
    \begin{center}
    \includegraphics[width=1.0\linewidth]{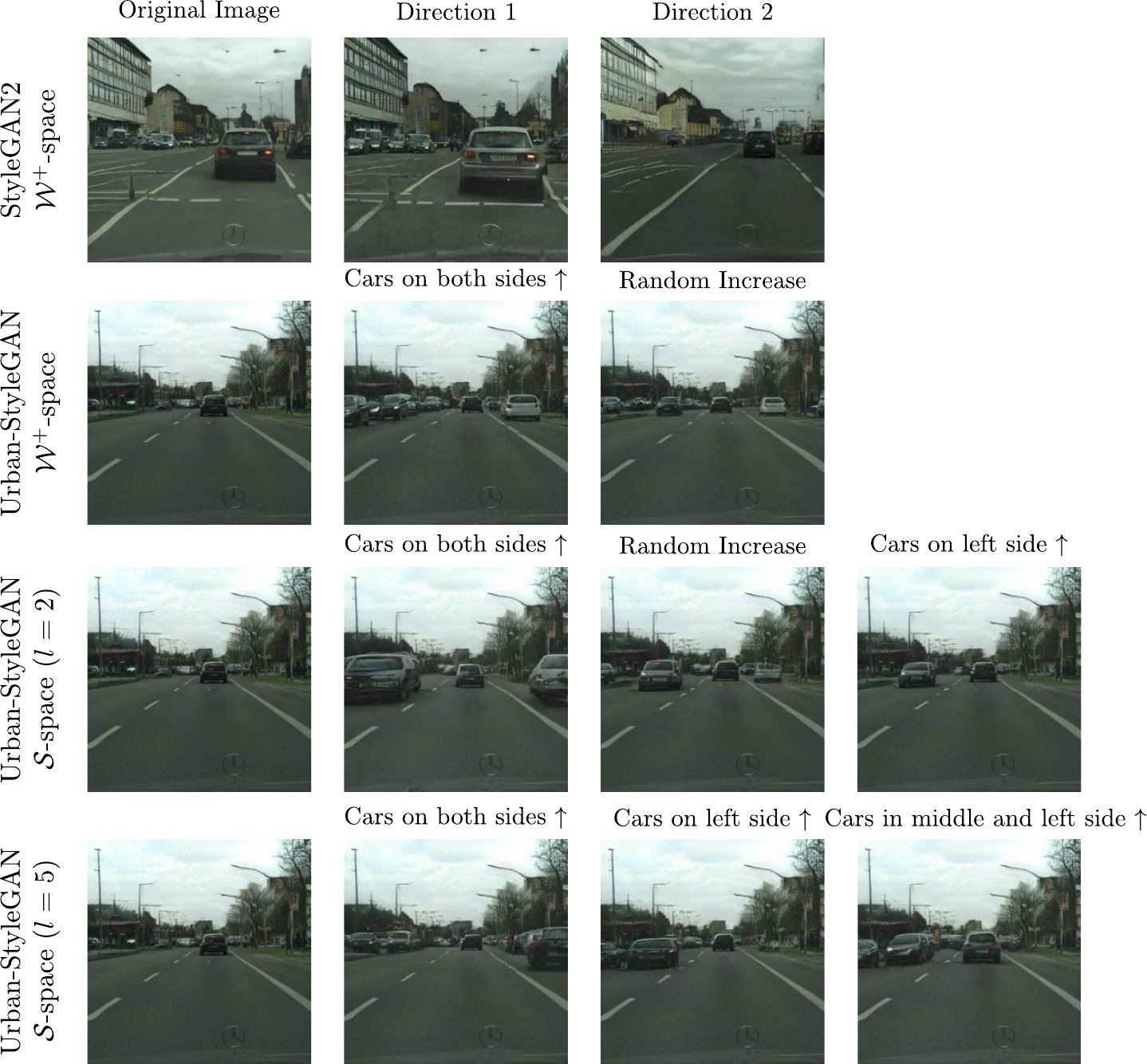}
    \end{center}
    \caption{Exploration of directions in different latent spaces. }
    \label{fig:tab2}
    \vspace{-1em}
\end{figure}
\begin{figure*}[t!]
    \centering
    \begin{center}
    \includegraphics[width=1.0\textwidth]{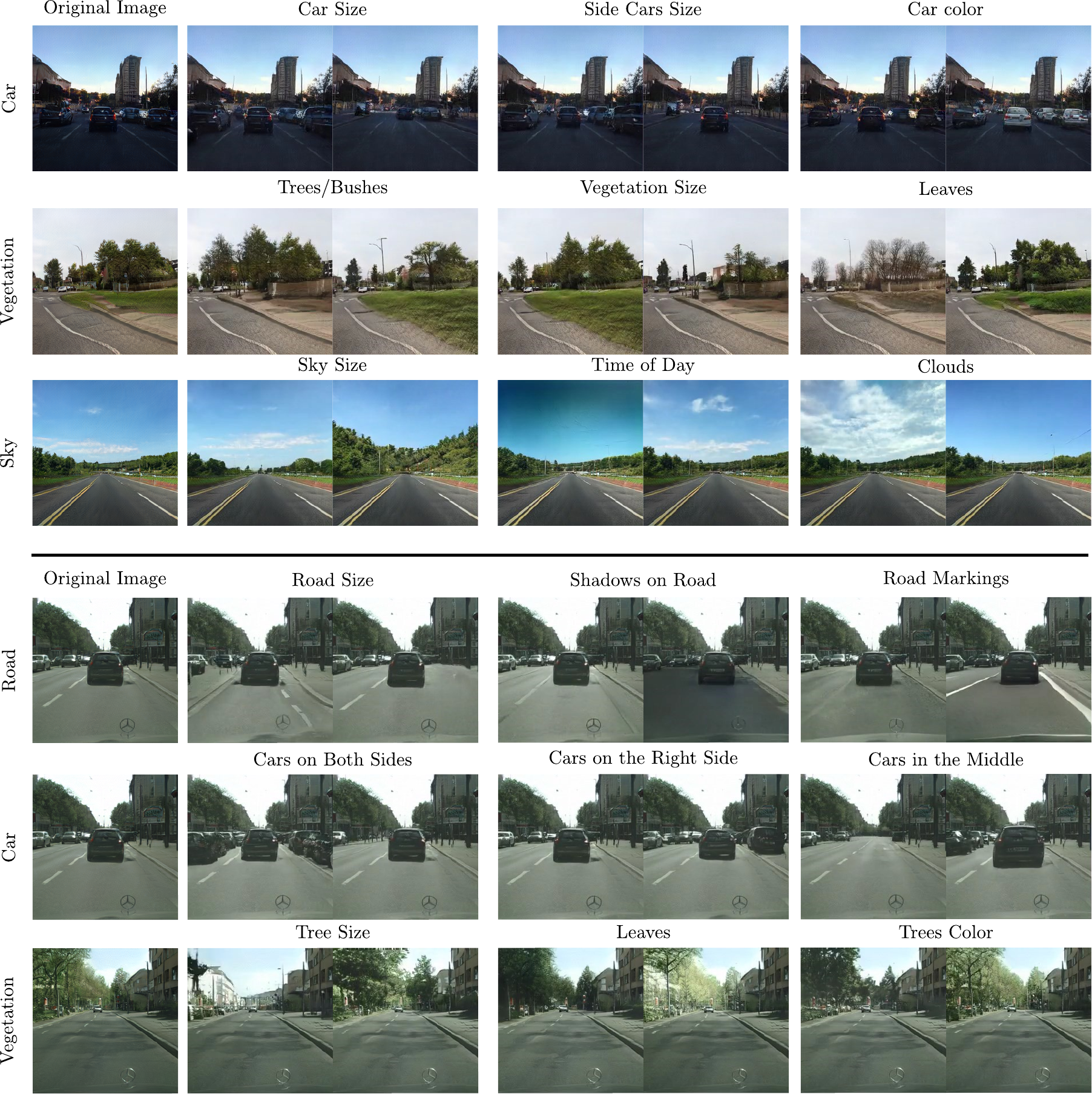}
    \end{center}
    \caption{More Results on Mapillary (upper part) and Cityscapes (lower part). In each row, we provide multiple directions of control for one class in the image. Two changes are shown for each direction ( example: Car size: big/small.) }
    \label{fig:tab3}
    \vspace{-1em}
\end{figure*}
\noindent\textbf{Main Comparisons} In Figure~\ref{fig:tab1}, we compare our approach with 2 baselines in terms of the quality and diversity of local edits. The first baseline is Semantic Palette~\cite{palette}, which offers controllability by explicitly increasing or decreasing the class distribution in the conditional input vector. We notice that while Semantic Palette is able to match the given conditional distribution, the effect of this manipulation is often unpredictable on other classes in the image. For instance, increasing the car distribution in row 1 Figure ~\ref{fig:tab1} leads to the appearance of more buildings and traffic signs, which should not be affected. Similarly, increasing the tree distribution leads to the car's disappearance on the left, although it is far from the trees' position in the original image. The second baseline is the native editing of SSG applied to our framework, which consists in increasing or decreasing the class latent vectors by a constant value. The edits look much more realistic and localized, proving that disentanglement in the latent code is key to controllability. However, we still observe minor inconsistencies, such as a slight change in the car's shape when the road changes, or the car in the middle becomes smaller while the cars on the side become larger. In contrast, our approach of applying PCA in the $\mathcal{S}$-space of the network shows more localized and meaningful edits in the image. 

\noindent\textbf{Which latent space should be manipulated?} In Figure~\ref{fig:tab2}, we show the merits of applying PCA in the $\mathcal{S}$-space of a disentangled latent space. For this, we first show the directions obtained by GANSpace~\cite{ganspace} applied to StyleGAN2. We note that no interpretable directions could be obtained, and the edits affect multiple parts of the image. Applying GANSpace in the $\mathcal{W}^+$-space of the class 'car' gives only one meaningful direction of control (increasing car number and size on both sides). On the other hand, the $\mathcal{S}$-space reveals more directions (increasing cars on one side or multiple specific parts of the image), with the late layer in the generator showing the largest number of meaningful edits. This confirms our assumption that the $\mathcal{S}$-space of the latest layers is the most disentangled.  

\noindent\textbf{Further qualitative results.} We show further qualitative results in Figure~\ref{fig:tab3} on multiple classes in Mapillary and Cityscapes. As Mapillary is more diverse than Cityscapes, more interesting directions, such as the sky size, cloudiness, and time of day, can be found. Moreover, we notice interesting directions that can be attributed to the effect of super-classes. The super-class 'vegetation' in Mapillary is the combination of the 'tree' and 'bush' classes, and we found one direction of control that increases 'trees' while decreasing 'bushes' in the image. Another direction can increase the 2 simultaneously. This shows that the latent space is more compact and disentangled.  

\section{Conclusion}
In this work, we investigate the effect of latent code disentanglement on generating urban scenes. We find that a disentangled latent code is vital for high-quality scene generation and that the key to exploiting disentanglement is to learn groups of classes together. This allows a more compact latent representation and regularizes the adversarial training by reducing the generator's learning capacity. We also find that the key to finding more disentangled directions of control is to explore the $\mathcal{S}$-space of the downstream class-specific layers. Our approach outperforms models for urban scene generation and is on par or slightly inferior to the state-of-the-art general-purpose models, which allow very little controllability. Our work is not without limitations. One drawback to our approach is that it needs label maps for training, which can be expensive and time-consuming. Future works can focus on how to provide a disentangled generation process in a totally unsupervised manner. On the other hand, we believe even more controllability can be achieved on the object level (as opposed to the class level in our framework) by using instance segmentation maps instead of semantic segmentation maps. While instance maps are more costly, they can lead to even more control over the generation process. Future directions include exploring diffusion models for synthesis or CLIP-based methods for editing with language prompts.

\label{sec:conclusion}

\clearpage
\clearpage
\begingroup
\printbibliography

@inproceedings{spade,
	title={Semantic image synthesis with spatially-adaptive normalization},
	author={Park, Taesung and Liu, Ming-Yu and Wang, Ting-Chun and Zhu, Jun-Yan},
	booktitle={Conference on Computer Vision and Pattern Recognition (CVPR)},
	year={2019}
}

@inproceedings{ccfpse,
	title={Learning to predict layout-to-image conditional convolutions for semantic image synthesis},
	author={Liu, Xihui and Yin, Guojun and Shao, Jing and Wang, Xiaogang and others},
	booktitle={Advances in Neural Information Processing Systems (NeurIPS)},
	year={2019}
}

@InProceedings{gtav,
author = {Stephan R. Richter and Vibhav Vineet and Stefan Roth and Vladlen Koltun},
title = {Playing for Data: {G}round Truth from Computer Games},
booktitle = {European Conference on Computer Vision (ECCV)},
year = {2016},
editor = {Bastian Leibe and Jiri Matas and Nicu Sebe and Max Welling},
series = {LNCS}, 
volume = {9906}, 
publisher = {Springer International Publishing},
pages = {102--118}
}

@inproceedings{pix2pixhd,
	title={High-resolution image synthesis and semantic manipulation with conditional {GANs}},
	author={Wang, Ting-Chun and Liu, Ming-Yu and Zhu, Jun-Yan and Tao, Andrew and Kautz, Jan and Catanzaro, Bryan},
	booktitle={Conference on Computer Vision and Pattern Recognition (CVPR)},
	year={2018}
}

@inproceedings{cordts2016cityscapes,
	title={The cityscapes dataset for semantic urban scene understanding},
	author={Cordts, Marius and Omran, Mohamed and Ramos, Sebastian and Rehfeld, Timo and Enzweiler, Markus and Benenson, Rodrigo and Franke, Uwe and Roth, Stefan and Schiele, Bernt},
	booktitle={Conference on Computer Vision and Pattern Recognition (CVPR)},
	year={2016}
}

@inproceedings{palette,
  title={Semantic Palette: Guiding Scene Generation with Class Proportions},
  author={Le Moing, Guillaume and Vu, Tuan-Hung and Jain, Himalaya and P{\'e}rez, Patrick and Cord, Matthieu},
  booktitle={Proceedings of the IEEE/CVF Conference on Computer Vision and Pattern Recognition},
  pages={9342--9350},
  year={2021}
}

@inproceedings{carla,
  title={CARLA: An open urban driving simulator},
  author={Dosovitskiy, Alexey and Ros, German and Codevilla, Felipe and Lopez, Antonio and Koltun, Vladlen},
  booktitle={Conference on robot learning},
  pages={1--16},
  year={2017},
  organization={PMLR}
}

@article{stylegan,
  title={A Style-Based Generator Architecture for Generative Adversarial Networks},
  author={Tero Karras and S. Laine and Timo Aila},
  journal={2019 IEEE/CVF Conference on Computer Vision and Pattern Recognition (CVPR)},
  year={2019},
  pages={4396-4405}
}

@article{stylegan2,
  title={Analyzing and Improving the Image Quality of StyleGAN},
  author={Tero Karras and S. Laine and M. Aittala and Janne Hellsten and J. Lehtinen and Timo Aila},
  journal={2020 IEEE/CVF Conference on Computer Vision and Pattern Recognition (CVPR)},
  year={2020},
  pages={8107-8116}
}

@inproceedings{stylegan3,
 author = {Karras, Tero and Aittala, Miika and Laine, Samuli and H\"{a}rk\"{o}nen, Erik and Hellsten, Janne and Lehtinen, Jaakko and Aila, Timo},
 booktitle = {Advances in Neural Information Processing Systems},
 editor = {M. Ranzato and A. Beygelzimer and Y. Dauphin and P.S. Liang and J. Wortman Vaughan},
 pages = {852--863},
 publisher = {Curran Associates, Inc.},
 title = {Alias-Free Generative Adversarial Networks},
 volume = {34},
 year = {2021}
}

@misc{Karras2018ProgressiveGO,
  title={Progressive Growing of GANs for Improved Quality, Stability, and Variation},
  author={Tero Karras and Timo Aila and S. Laine and J. Lehtinen},
  journal={ArXiv},
  year={2018},
  howpublished = "\url{http://arxiv.org/abs/1710.10196}"
}

@inproceedings{
oasis,
title={You Only Need Adversarial Supervision for Semantic Image Synthesis},
author={Edgar Sch{\"o}nfeld and Vadim Sushko and Dan Zhang and Juergen Gall and Bernt Schiele and Anna Khoreva},
booktitle={International Conference on Learning Representations},
year={2021}
}

@inproceedings{Goodfellow2014GenerativeAN,
  title={Generative Adversarial Nets},
  author={I. Goodfellow and Jean Pouget-Abadie and Mehdi Mirza and Bing Xu and David Warde-Farley and Sherjil Ozair and Aaron C. Courville and Yoshua Bengio},
  booktitle={NIPS},
  year={2014}
}

@article{usiscag,
title = {USIS: Unsupervised Semantic Image Synthesis},
journal = {Computers \& Graphics},
year = {2023},
issn = {0097-8493},
doi = {https://doi.org/10.1016/j.cag.2022.12.010},
url = {https://www.sciencedirect.com/science/article/pii/S0097849323000018},
author = {George Eskandar and Mohamed Abdelsamad and Karim Armanious and Bin Yang},
}

@inproceedings{ssg,
  title={SemanticStyleGAN: Learning Compositional Generative Priors for Controllable Image Synthesis and Editing},
  author={Shi, Yichun and Yang, Xiao and Wan, Yangyue and Shen, Xiaohui},
  booktitle={Proceedings of the IEEE/CVF Conference on Computer Vision and Pattern Recognition},
  pages={11254--11264},
  year={2022}
}

@inproceedings{mapillary,
  title={The mapillary vistas dataset for semantic understanding of street scenes},
  author={Neuhold, Gerhard and Ollmann, Tobias and Rota Bulo, Samuel and Kontschieder, Peter},
  booktitle={Proceedings of the IEEE international conference on computer vision},
  pages={4990--4999},
  year={2017}
}

@article{sbgan,
  title={Semantic bottleneck scene generation},
  author={Azadi, Samaneh and Tschannen, Michael and Tzeng, Eric and Gelly, Sylvain and Darrell, Trevor and Lucic, Mario},
  journal={arXiv preprint arXiv:1911.11357},
  year={2019}
}

@inproceedings{vqgan,
  title={Taming transformers for high-resolution image synthesis},
  author={Esser, Patrick and Rombach, Robin and Ommer, Bjorn},
  booktitle={Proceedings of the IEEE/CVF conference on computer vision and pattern recognition},
  pages={12873--12883},
  year={2021}
}

@article{projected,
  title={Projected gans converge faster},
  author={Sauer, Axel and Chitta, Kashyap and M{\"u}ller, Jens and Geiger, Andreas},
  journal={Advances in Neural Information Processing Systems},
  volume={34},
  pages={17480--17492},
  year={2021}
}

@inproceedings{styleganxl,
  title={Stylegan-xl: Scaling stylegan to large diverse datasets},
  author={Sauer, Axel and Schwarz, Katja and Geiger, Andreas},
  booktitle={ACM SIGGRAPH 2022 conference proceedings},
  pages={1--10},
  year={2022}
}

@article{ganspace,
  title={Ganspace: Discovering interpretable gan controls},
  author={H{\"a}rk{\"o}nen, Erik and Hertzmann, Aaron and Lehtinen, Jaakko and Paris, Sylvain},
  journal={Advances in Neural Information Processing Systems},
  volume={33},
  pages={9841--9850},
  year={2020}
}

@inproceedings{sefa,
  title={Closed-form factorization of latent semantics in gans},
  author={Shen, Yujun and Zhou, Bolei},
  booktitle={Proceedings of the IEEE/CVF conference on computer vision and pattern recognition},
  pages={1532--1540},
  year={2021}
}

@inproceedings{fastgan,
  title={Towards faster and stabilized gan training for high-fidelity few-shot image synthesis},
  author={Liu, Bingchen and Zhu, Yizhe and Song, Kunpeng and Elgammal, Ahmed},
  booktitle={International Conference on Learning Representations},
  year={2021}
}

@inproceedings{stylemc,
  title={StyleMC: multi-channel based fast text-guided image generation and manipulation},
  author={Kocasari, Umut and Dirik, Alara and Tiftikci, Mert and Yanardag, Pinar},
  booktitle={Proceedings of the IEEE/CVF Winter Conference on Applications of Computer Vision},
  pages={895--904},
  year={2022}
}

@inproceedings{styleclip,
  title={Styleclip: Text-driven manipulation of stylegan imagery},
  author={Patashnik, Or and Wu, Zongze and Shechtman, Eli and Cohen-Or, Daniel and Lischinski, Dani},
  booktitle={Proceedings of the IEEE/CVF International Conference on Computer Vision},
  pages={2085--2094},
  year={2021}
}

@inproceedings{discofacegan,
  title={Disentangled and controllable face image generation via 3d imitative-contrastive learning},
  author={Deng, Yu and Yang, Jiaolong and Chen, Dong and Wen, Fang and Tong, Xin},
  booktitle={Proceedings of the IEEE/CVF conference on computer vision and pattern recognition},
  pages={5154--5163},
  year={2020}
}

@inproceedings{config,
  title={Config: Controllable neural face image generation},
  author={Kowalski, Marek and Garbin, Stephan J and Estellers, Virginia and Baltru{\v{s}}aitis, Tadas and Johnson, Matthew and Shotton, Jamie},
  booktitle={Computer Vision--ECCV 2020: 16th European Conference, Glasgow, UK, August 23--28, 2020, Proceedings, Part XI 16},
  pages={299--315},
  year={2020},
  organization={Springer}
}

@inproceedings{gancontrol,
  title={Gan-control: Explicitly controllable gans},
  author={Shoshan, Alon and Bhonker, Nadav and Kviatkovsky, Igor and Medioni, Gerard},
  booktitle={Proceedings of the IEEE/CVF international conference on computer vision},
  pages={14083--14093},
  year={2021}
}

@article{spectralnorm,
  title={Spectral normalization for generative adversarial networks},
  author={Miyato, Takeru and Kataoka, Toshiki and Koyama, Masanori and Yoshida, Yuichi},
  journal={arXiv preprint arXiv:1802.05957},
  year={2018}
}

@article{styleflow,
  title={Styleflow: Attribute-conditioned exploration of stylegan-generated images using conditional continuous normalizing flows},
  author={Abdal, Rameen and Zhu, Peihao and Mitra, Niloy J and Wonka, Peter},
  journal={ACM Transactions on Graphics (ToG)},
  volume={40},
  number={3},
  pages={1--21},
  year={2021},
  publisher={ACM New York, NY}
}

@inproceedings{interfacegan,
  title={Interpreting the latent space of gans for semantic face editing},
  author={Shen, Yujun and Gu, Jinjin and Tang, Xiaoou and Zhou, Bolei},
  booktitle={Proceedings of the IEEE/CVF conference on computer vision and pattern recognition},
  pages={9243--9252},
  year={2020}
}

@inproceedings{sagan,
  title={Self-attention generative adversarial networks},
  author={Zhang, Han and Goodfellow, Ian and Metaxas, Dimitris and Odena, Augustus},
  booktitle={International conference on machine learning},
  pages={7354--7363},
  year={2019},
  organization={PMLR}
}

@article{segmentor,
  title={Rethinking atrous convolution for semantic image segmentation},
  author={Chen, Liang-Chieh and Papandreou, George and Schroff, Florian and Adam, Hartwig},
  journal={arXiv preprint arXiv:1706.05587},
  year={2017}
}
\endgroup

\end{document}